\documentclass{article}

\usepackage{microtype}
\usepackage{graphicx}
\usepackage{subfigure}
\usepackage{booktabs} 
\usepackage{multirow}

\usepackage{hyperref}

\usepackage[accepted]{icml2023}

\usepackage{amsmath}
\usepackage{amssymb}
\usepackage{mathtools}
\usepackage{amsthm}

\usepackage[capitalize,noabbrev]{cleveref}

\theoremstyle{plain}

\theoremstyle{definition}

\theoremstyle{remark}

\usepackage[textsize=tiny]{todonotes}

\icmltitlerunning{Ske2Grid: Skeleton-to-Grid Representation Learning for Action Recognition}

\begin{document}

\twocolumn[
\icmltitle{Ske2Grid: Skeleton-to-Grid Representation Learning for Action Recognition}



\icmlsetsymbol{equal}{*}

\begin{icmlauthorlist}
\icmlauthor{Dongqi Cai}{comp}
\icmlauthor{Yangyuxuan Kang}{comp}
\icmlauthor{Anbang Yao}{comp}
\icmlauthor{Yurong Chen}{comp}
\end{icmlauthorlist}

\icmlaffiliation{comp}{Intel Labs China}

\icmlcorrespondingauthor{Anbang Yao}{anbang.yao@intel.com}

\icmlkeywords{Skeleton-based Action Recognition, Graph Convolution Networks, Grid Patch}

\vskip 0.3in
]



\printAffiliationsAndNotice{}  

\begin{abstract}
This paper presents Ske2Grid, a new representation learning framework for improved skeleton-based action recognition.
In Ske2Grid, we define a regular convolution operation upon a novel grid representation of human skeleton, which is a compact image-like grid patch constructed and learned through three novel designs.
Specifically, we propose a graph-node index transform (GIT) to construct a regular grid patch through assigning the nodes in the skeleton graph one by one to the desired grid cells.
To ensure that GIT is a bijection and enrich the expressiveness of the grid representation, an up-sampling transform (UPT) is learned to interpolate the skeleton graph nodes for filling the grid patch to the full.
To resolve the problem when the one-step UPT is aggressive and further exploit the representation capability of the grid patch with increasing spatial size, a progressive learning strategy (PLS) is proposed which decouples the UPT into multiple steps and aligns them to multiple paired GITs through a compact cascaded design learned progressively.
We construct networks upon prevailing graph convolution networks and conduct experiments on six mainstream skeleton-based action recognition datasets.
Experiments show that our Ske2Grid significantly outperforms existing GCN-based solutions under different benchmark settings, without bells and whistles.
Code and models are available at \url{https://github.com/OSVAI/Ske2Grid}.
\end{abstract}

\section{Introduction}
\label{introduction}

\begin{figure}[ht]
\vskip 0.in
\begin{center}
\centerline{\includegraphics[width=\columnwidth]{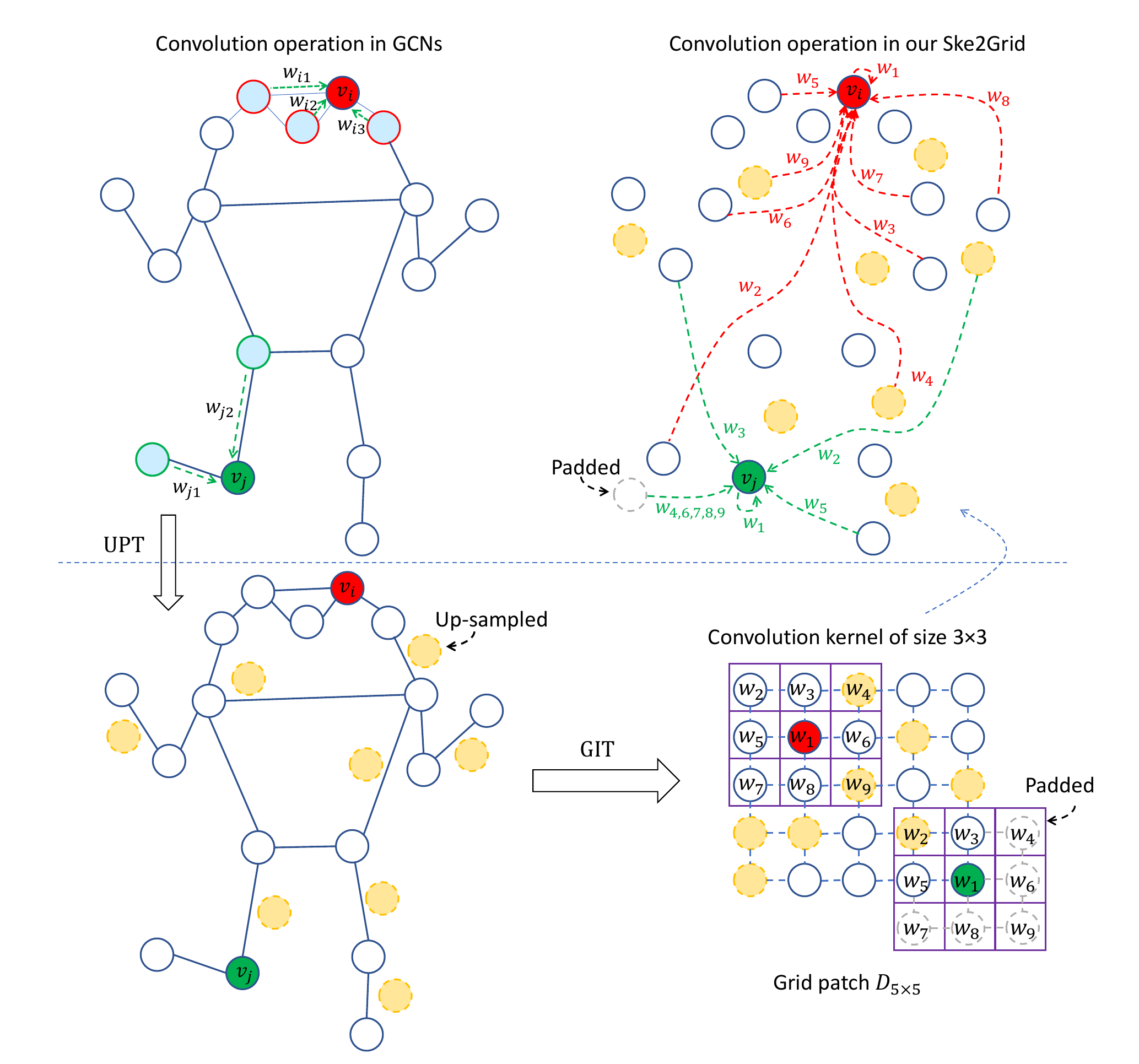}}
\caption{Comparison of convolution operations in GCNs and in our Ske2Grid. Graph convolution (top-left) typically convolves every node with their neighboring nodes using specific kernels. In Ske2Grid, we construct a regular grid patch for skeleton representation via up-sampling transform (UPT) and graph-node index transform (GIT). Convolution operation upon this grid patch convolves every grid cell using a shared regular kernel. It operates on a set of grid cells within a squared sub-patch which may be filled by a set of nodes distributed remotely on the graph, achieving a learnable receptive field on the skeleton for action feature modeling (top-right). In the figure, the up-sampled skeleton graph is visualized assuming the locations of the original graph nodes being unchanged for a better illustration.}
\label{ske2grid_overview}
\end{center}
\vskip -0.2in
\end{figure}

With the rapid development of 3D motion capturing systems and advanced real-time 2D/3D pose estimation algorithms, skeleton-based action recognition has attracted increasing attention from both industry and academia.
The performance of a skeleton-based action recognition system depends on how well it can model the discriminative skeleton feature interactions among the active coordinated human joints when performing an action.
Traditional methods \cite{Ref07_Handcraft_CNN,Ref07_Handcraft_FreqAttention,Ref07_Handcraft_FuseHighOrder} focus on designing hand-crafted features to model spatial structure and temporal dynamics of skeleton sequences, and use well-designed classifiers to recognize human actions.
Early deep learning based methods \cite{Ref01_RNN_Hierarchical_RNN,Ref01_RNN_ST_LSTM,Ref01_RNN_IndepRNN} consider skeletons in videos as temporal vector sequences and use deep recurrent networks such as RNN and LSTM to model the skeleton dynamics.
In recent years, convolutional neural networks (CNNs) and graph convolution networks (GCNs) have become the mainstream learning models in skeleton-based action recognition research.
CNN-based methods \cite{Ref02_CNN_visualization,Ref02_CNN_pa3d,Ref02_CNN_PoseC3D} usually convert skeleton to image-like input and use 2D or 3D CNNs for end-to-end feature extraction.
GCN-based methods \cite{Ref03_GCN_ST_GCN,Ref03_GCN_AS_GCN,Ref03_GCN_GCN_NAS} treat skeleton as an irregular graph and learn to aggregate the skeleton features 
in terms of the pre-defined topology of graph.
More recently, hypergraph-based models, such as \citet{HyperGNN}, also adopt a skeleton graph structure. Unlike conventional graph-based models that usually ignore non-physical dependencies among joints of the human body, hypergraph-based models address this gap by introducing local and global hyperedges to encode higher-order feature dependencies.
Besides, newly emerging transformer-based models, such as \citet{STST}, \citet{ST-TR} and \citet{Hyperformer}, still retain a skeleton graph structure and treat each joint of the human body as a token, and then use spatial and temporal self-attention operators to capture feature dependencies.

Despite the prevalence of CNNs and GCNs based solutions, improved representation learning for skeleton-based action recognition is still a challenging problem.
On the one side, CNN-based solutions benefit from the regular convolution for efficient feature modeling. However, they are restricted to image-like input, to which the conversion from skeleton typically ignores the critical topological information of human joints and would sacrifice the compactness of skeleton data to some extent.
For example, \citet{Ref07_Handcraft_image_input_mtdl} directly encodes the temporal dimension, the joints and the coordinates as different axis of the image-like tensor.
In the recent work of \citet{Ref02_CNN_PoseC3D}, skeleton is represented as the stacked heatmaps of joints obtained by 2D pose estimators.
Besides, the image-like input may then be fixed during training, reducing the flexibility of skeleton in representing action patterns.
On the other side, GCN-based solutions benefit from the utilization of topological relations among skeleton joints for feature modeling. However, they rely on the irregular topological structure and are forced to learn different shaped and separate convolution kernels node by node, lacking similar properties of feature aggregation in formulation.
Furthermore, the receptive field of graph convolution usually covers a set of adjacent nodes with pre-defined distances to the target node, degrading the efficiency for modeling feature interactions especially among human joints that are actively coordinated yet topologically remote.

Motivated by these two observations, we formulate a new skeleton representation called Ske2Grid for action recognition with the expectation to inherit the non-Euclidean spatial layout of the graph skeleton while enabling the use of regular convolution as in the Euclidean image space but with a compact shape.
Specifically, we investigate the problem from two technical perspectives.
(1) from representation perspective, how to convert the skeleton from an irregular graph to a regular image-like grid patch while maintaining its critical topological representation capability and compactness for action recognition?
(2) from network perspective, how to guarantee that the learned grid patch has improved representation abilities over the skeleton graph for action representation?
To the first question, we design a novel grid representation of skeleton constructed and learned through three novel designs.
Specifically, we construct a regular grid patch through allocating the nodes in the skeleton graph one by one to the desired grid cells using a graph-node index transform (GIT), inspired by the recent work \cite{GridConv} for 3D human pose estimation.
To ensure that GIT is a bijection and enrich the expressiveness of the grid representation, we propose an up-sampling transform (UPT) with regulation of the graph topology to interpolate the skeleton graph nodes for filling the grid patch to the full.
To further exploit the representation capability of the grid patch with increasing spatial size under which the one-step UPT is aggressive and tends to lose its effectiveness, a progressive learning strategy (PLS) is proposed for decoupling the UPT into multiple steps and aligning them to multiple paired GITs through a compact cascaded design learned progressively.
The resulting grid representation differs from the existing image-like skeleton inputs in the following ways:
(1) the grid patch is filled by up-sampled graph nodes with its layout reflecting the topological relations of skeleton features for action modeling;
(2) the grid patch representation is much more compact in size compared with images (e.g. $5\!\times\!5\!\times\!3$). Instead of being down-sampled as in CNNs, the spatial size of our grid patch is fixed during representation learning to maintain the compact skeleton semantics;
(3) the GIT and UPT can be jointly learned with the Ske2Grid convolution network in an end-to-end manner, enabling the enriched topological connections on the skeleton being automatically learned conditioned on the target action recognition dataset.
To the second question, we define a regular convolution operation upon our grid patch for efficiently modeling the skeleton feature interactions.
As shown in \cref{ske2grid_overview}, with the above novel designs for Ske2Grid, we can readily use the regular convolution with the square-shaped convolution kernel to learn discriminative feature interactions by naturally sharing the kernel to all grid nodes.
Benefiting from the cascaded structure of multiple UPT and GIT pairs facilitated by the PLS, the resulting grid patch incorporates extra nodes interpolated from the original graph nodes, which further enables the regular convolution to capture various ordered topological relations, significantly strengthening the learning ability of the regular convolution on the grid patch.

Our Ske2Grid could be readily applicable to popular GCN architectures for improved skeleton-based action recognition, without modifying their built-in temporal modules.
We construct networks through replacing spatial graph convolution in prevailing GCNs with regular 2D convolution upon our grid patch representation, and conduct experiments on six mainstream skeleton-based action recognition datasets.
Experiments show that our method significantly outperforms many existing GCN-based solutions under different settings, demonstrating the effectiveness of our method in learning improved skeleton features for action recognition.

\section{Related Works}
\label{related_works}

\textbf{Skeleton-based action recognition with CNNs.} Inspired by the research progress in image-based tasks, it is natural to convert the graph-structured skeleton into regular image-like input and take the advantage of image recognition pipelines to handle skeleton-based action recognition task. This line of works obtains a sequence of 2D array via combining skeleton features spatially and temporally, and then directly transforms the 2D arrays to gray images.
For example, \citet{Ref07_Handcraft_CNN} transforms skeleton into three clips consisting of several frames represented by computed spatial temporal features, and uses a deep CNN to model long-term temporal dynamics of skeleton.
\citet{Ref07_Handcraft_image_input_mtdl} directly encodes the temporal dimension, the joints and the coordinates to be different axis of a pseudo image, and uses a CNN-based multitask framework for pose estimation and action recognition. \citet{Ref02_CNN_PoseC3D} takes the stacked heatmaps of skeleton joints obtained by 2D pose estimators as input, and utilizes 3D convolutional neural networks to recognize actions, which refreshes the state-of-the-art on many skeleton-based action recognition benchmarks.
These solutions benefit from the regular 2D/3D convolution for efficient image feature modeling. However, the conversion from skeleton to image-like input usually ignores the critical topological information of human joints and would sacrifice the compactness of skeleton data to some extent. Besides, the image-like input is usually fixed during training.
Unlike them, we construct and learn a compact grid representation for skeleton, and its layout reflects the topological relations of human joints for action modeling.

\textbf{Skeleton-based action recognition with GCNs.} Skeleton can be naturally represented by an irregular graph, in which nodes represent joint coordinates and edges naturally connect joints in human bodies. GCNs generalize CNNs to graphs of arbitrary structures which are mainstream learning models to handle graph-structured skeleton data.
ST-GCN \cite{Ref03_GCN_ST_GCN} extends GCNs to a spatial-temporal graph model and designs convolution kernels for skeleton modeling, which is the first work that achieves satisfactory performance on large-scale skeleton-based action recognition benchmarks.
However, it uses a pre-defined topology according to human body structure, which is fixed during both training and testing phases.
A lot of following methods improve ST-GCN through constructing skeleton graph with dynamic topologies \cite{Ref03_GCN_DG_STGCN,Ref03_GCN_Two_Stream_AGCN,Ref03_GCN_MS_AAGCN}.
Many recent methods improve the receptive field of graph convolution through incorporating extra contextual information.
AS-GCN \cite{Ref03_GCN_AS_GCN} introduces an A-link module to capture action-specific latent dependencies and an S-link module to represent higher-order node dependencies of skeleton graph.
CA-GCN \cite{Ref03_GCN_Context_Aware_GCN} introduces a context term for each vertex by integrating information from the entire skeleton graph to enlarge the receptive field of graph convolution.
Shift-GCN \cite{Ref03_GCN_Shift_GCN} proposes local shift graph operation and non-local shift graph operation to provide flexible receptive fields for spatial and temporal graphs.
CTR-GCN \cite{Ref03_GCN_CTR-GCN} proposes to learn a shared topology and channel-specific correlations simultaneously, obtaining channel-wise topologies which improve feature aggregations in graph convolution.
Despite the advancement of these solutions, they still model the irregular skeleton graph using standard graph convolution.
As discussed in the previous section, we define a regular convolution operation upon our learnable grid patch for efficiently modeling the skeleton feature interactions.

\section{Method}
\label{method}

In this section, we first revisit the graph convolution from a general perspective, then define the convolution operation upon our grid patch along with three core designs in Ske2Grid, and finally introduce the construction of network.

\begin{figure*}[ht]
\vskip 0.in
\begin{center}
\centerline{\includegraphics[scale=0.45]{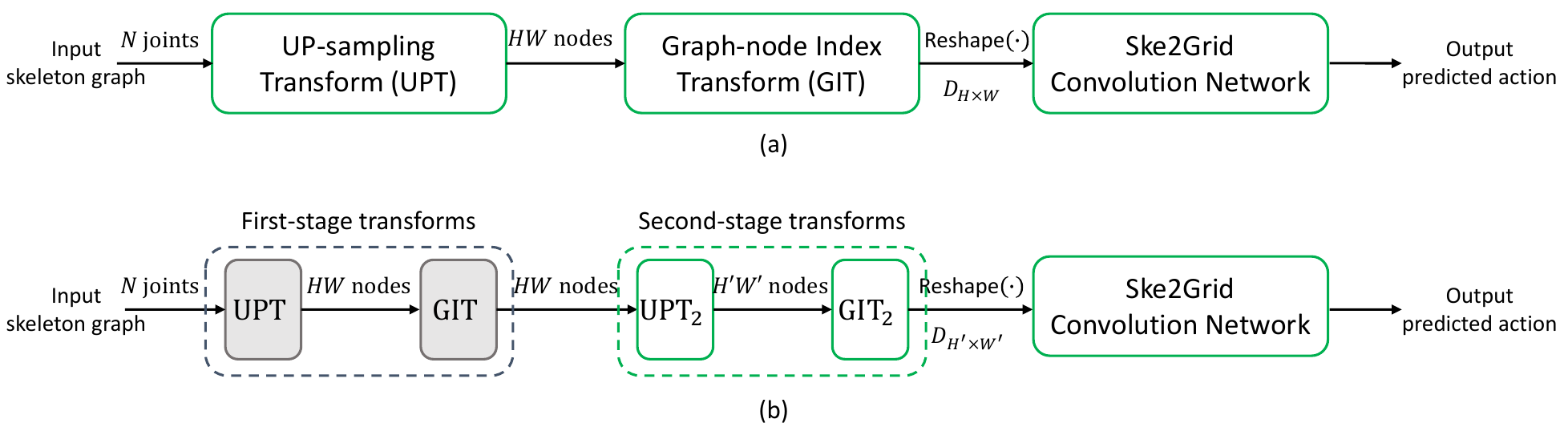}}
\caption{(a) The overall framework of Ske2Grid: the input skeleton graph with $N$ joints is converted to a grid patch of size ${H\!{\times}\!{W}}$ using a pair of up-sampling transform (UPT) and graph-node index transform (GIT), which is then fed into the Ske2Grid convolution network for action recognition. (b) Ske2Grid with progressive learning strategy (PLS): the input skeleton is converted to a larger grid patch ($H\!'>\!H, W\!'>\!W$) using two-stage UPT plus GIT pairs. The well-trained Ske2Grid convolution network for the first-stage grid patch as in (a) is re-used to initialize the network for the second-stage grid patch as in (b), and the first-stage UPT plus GIT pair is fixed during training. PLS is used in a cascaded way to boost the performance of our Ske2Grid convolution network with increasing grid patch size. }
\label{ske2grid_pipeline}
\end{center}
\vskip -0.3in
\end{figure*}

\subsection{Graph Convolution}
\label{graph_conv}

Considering a skeleton graph $G=\{V,E,A\}$, which consists of a set of nodes $V$ with $|V|=N$, a set of edges $E$ with $|E|=M$ and the adjacency matrix $\boldsymbol{A}\in\{0,1\}^{N\times N}$.
If there is an edge between nodes $v_i$ and $v_j$, the entry $\boldsymbol{A}(i,j)\!=\!1$; otherwise, $\boldsymbol{A}(i,j)\!=\!0$.
We represent the corresponding feature of the graph as $\boldsymbol{X}\in{\mathbb{R}^{N\times{C}}}$, where $C$ denotes the skeleton feature dimension, and $\boldsymbol{x}_i\in{\mathbb{R}^{C\times 1}}$ denotes the feature vector for node $i$.
The output value of graph convolution operation for a single channel at the node $i$ can be written as

\begin{equation}{\label{eq_graphconv}}
f_{out}(v_i)=\sum_{v_j\in{B(v_i)}}\boldsymbol{w}_{i,j}{\boldsymbol{x}_j},
\end{equation}

where $B(v_i)\!=\!\{v_j|d(v_j,v_i)\leq{D}\}$ denotes the neighbor set of node $i$; $d(v_j,v_i)$ is the minimum length of any path from node $j$ to node $i$, and normally $D$ is set to $1$;
$\boldsymbol{w}_{i,j}\in{\mathbb{R}^{1\times C}}$ is a weight vector specific to node $i$ for computing the inner product with the corresponding input feature $\boldsymbol{x}_j$, which is indexed within the neighbor set $B(v_i)$.
This convolution operation models the feature interactions among a set of neighboring nodes defined by the graph topology.
However, the improved expressiveness of the skeleton action representation lies in its capability of modeling the discriminative feature interactions among a set of actively coordinated joints when performing an action, which may require the neighbor set activated in convolution operation to be flexible and expressive.

\subsection{Convolution in Ske2Grid}
\label{ske2gridconv}

It is intuitive to make the neighbor set in graph convolution be learnable to achieve improved expressiveness for action representation. However, the convolution kernel is specific to the target node and is difficult to model the correlations among a changing set of nodes.
To solve this problem, we define a regular convolution operation in Ske2Grid.

Inspired by the recent progress in 3D human pose estimation \cite{GridConv}, we construct a regular grid patch $D_{H\!\times\!W}$ with $H$ and $W$ being the spatial height and width, respectively.
It consists of a set of grid cells $d_{i,j}$ filled by specific nodes from the skeleton graph.
The feature of the grid patch is denoted as $\boldsymbol{Y}\in{\mathbb{R}^{H\!\times\! W\!\times{C}}}$, and $\boldsymbol{y}_{i,j}\in \mathbb{R}^{C\times 1}$ is the feature vector at the spatial location $(i,j)$.
Denote the size of a regular convolution kernel as $K\!\times\! K$.
Similar to \cref{eq_graphconv}, the output value of our convolution operation upon a grid cell at the spatial location $(i,j)$ for a single channel is defined as

\begin{equation}{\label{eq_skeconv}}
f_{out}(d_{i,j})=\sum_{d_{m,n}\in{B_D({d_{i,j}})}}\boldsymbol{w}_{k}{\boldsymbol{y}_{m,n}},
\end{equation}

where $B_D(d_{i,j})$ is the neighbor set of the grid cell $d_{i,j}$, which is the $K\!\times\!K$ sub-patch centered on $d_{i,j}$ containing $K^2$ grid cells; the set $\{\boldsymbol{w}_{k}\in \mathbb{R}^{1\times C}|k\!=\!1,...,K^2\}$ constitutes a squared-shaped convolution kernel for computing the inner product with features in the neighbor set of the grid cell $d_{i,j}$ 
as shown in \cref{ske2grid_overview}.
In sharp contrast to graph convolution, the square-shaped convolution kernel in our Ske2Grid is shared everywhere on the grid patch, which is key to model a learnable set of grid cells within a regular neighbor set. Now, the question is how to learn a decent layout of the grid patch, making the neighboring grid cells be expressive for action representation.


\subsection{Graph-node Index Transform}
\label{git}

As mentioned above, a grid patch is constructed through assigning the nodes in the graph one by one to the grid cells.
The problem is how to fill each grid cell with an appropriate graph node for improved feature interaction modeling.
We propose to learn a mapping from the indexes of nodes in $G$ to the spatial indexes of grid cells in $D_{H\!\times\!W}$, which is called graph-node index transform (GIT).

The GIT from $N$ graph nodes to the grid patch $D_{H\!\times\! W}$ is defined by a binary matrix $\boldsymbol{\Phi}\! \in\!{\{0,1\}^{HW\times{N}}}$ , in which each row $\boldsymbol{\phi}_i\!\in\! \{0,1\}^{N}$ is a one-hot vector indicating the index of the one selected node in the graph. That is, a grid cell $d_i$ is filled with a specific graph node $v_j$ only if $\boldsymbol{\phi}_{i,j}\!=1$.
With this definition, the feature of the grid patch $D_{H\!\times\! W}$ can be obtained by

\begin{equation}{\label{eq_GIT}}
\boldsymbol{Y} = \rm {reshape}(\boldsymbol{\Phi} \cdot \boldsymbol{X}),
\end{equation}

where ``$\cdot$'' is a row-by-column multiplication, with each row of the product matrix being the selected graph node feature.
The $\rm {reshape}(\cdot)$ operation rearranges the output of $\boldsymbol{\Phi}\!\cdot\!\boldsymbol{X}$ into an $H\!\times\!W$ grid patch representation.
Thus, the layout of the grid patch can be learned along with $\boldsymbol{\Phi}$.

However, directly learning a binary matrix will cut off the backward gradient flow, which makes the training non-differentiable.
To address this problem, we use straight-through estimator (STE) \cite{Ref08_Others_STE} for parameter update.
Specifically, we introduce a real-value matrix $\boldsymbol{\Psi}\in{\mathbb{R}^{HW\!\times\! N}}$ to assist the learning of $\boldsymbol{\Phi}$.
During training, we obtain $\boldsymbol{\Phi}$ through binarizing $\boldsymbol{\Psi}$ row by row according to

\begin{equation}{\label{eq_binary}}
\boldsymbol{\phi}_{i,j} = \begin{cases}
1 &\text{if\ } \boldsymbol{\psi}_{i,j}=\mathop{\max}(\boldsymbol{\psi}_{i,\cdot}),\\
0 &\text{otherwise.}
\end{cases}
\end{equation}

Specifically, only one element in each row of $\boldsymbol{\Phi}$ is set to one, at its position where the maximum value in the corresponding row of $\boldsymbol{\Psi}$ appears.
In the backward, continuous gradient is used to update the real-value matrix $\boldsymbol{\Psi}$, instead of $\boldsymbol{\Phi}$.
In principle, by introducing $\boldsymbol{\Psi}$ as a continuous approximation of $\boldsymbol{\Phi}$, it enables the searching for a decent layout of grid patch for improved action expressiveness.

\subsection{Up-sampling Transform}
\label{upsamp}

To avoid information loss during the conversion from skeleton to grid representation, the number of grid cells should be set to no less than the number of skeleton joints.
Then all the skeleton joints can be assigned to the grid patch through making the GIT a surjection. 
However, the number of grid cells in the target grid patch is typically larger than the number of graph nodes, which means there exist some extra grid cells need to be filled.
When filling these grid cells with duplicated graph nodes, such an optimization problem is not trivial, potentially degrading the expressiveness of the grid representation.
To further guarantee that GIT is a bijection, we introduce an up-sampling transform (UPT) to interpolate skeleton graph nodes to fill the grid patch to the full.
The feature map of the up-sampled graph is obtained by

\begin{equation}{\label{eq_UPT_basic}}
\boldsymbol{X}' = \boldsymbol{\Lambda} \cdot \boldsymbol{X},
\end{equation}

where $\boldsymbol{\Lambda} \!\in\!{\mathbb{R}^{HW\!\times\!{N}}}$ denotes the up-sampling matrix and $\boldsymbol{X}'\!\in\! \mathbb{R}^{HW\!\times\!{C}}$ is the up-sampled feature.
Considering that the original topology of skeleton graph reveals intuitive coordinated movements of joints when performing an action, we further propose to incorporate the prior adjacency matrix into \cref{eq_UPT_basic} to regulate the up-sampling process

\begin{equation}{\label{eq_UPT}}
\boldsymbol{X}' = (\boldsymbol{\Lambda} \cdot \boldsymbol{A}) \cdot \boldsymbol{X}.
\end{equation}

The utilization of the adjacency matrix $\boldsymbol{A}$ encourages the UPT to interpolate new nodes using adjacent nodes along existing edges, which facilitates our skeleton-to-grid representation with the association of topological priors for improved action representation.

After the UPT, our grid representation can be obtained using \cref{eq_GIT} with the up-sampled feature and one-to-one index mapping.
In principle, the main idea behind the UPT is learning to interpolate the skeleton graph nodes for the improved expressiveness meanwhile ensuring that the following GIT for constructing our grid patch is a bijection.
Generally, GIT can be made as a bijection in various ways.
In our case, we simply add a non-repetitive constraint to the binarization process in \cref{eq_binary} through binarizing $\boldsymbol{\Psi}$ row by row with a simple greedy search.

\subsection{Progressive Learning Strategy}
\label{progressive}

With the GIT and UPT, the skeleton input can be converted into a grid patch of any size.
In image-based tasks, a CNN model tends to show improved performance when increasing the resolution of the input.
However, the performance improvement is marginal when directly increasing the size of grid patch in our method.
We conjecture that there exists a mismatch between the one-step transform pair of UPT and GIT and the straightforward learning strategy, leading to slightly worse performance than the original skeleton graph when the one-step UPT is aggressive.

To solve this problem, we propose progressive learning strategy (PLS), a novel optimization scheme, which decouples the aggressive one-step UPT into multiple steps and aligns them to multiple paired GITs through a compact cascaded design learned in a progressive manner.
In this way, the grid representation is gradually enriched through increasing the size of grid patch progressively.

Specifically, we first learn to convert the skeleton graph to a base grid patch $D_{H\!\times\! W}$.
Then, the previous-stage transform pair of UPT and GIT for the base grid patch is reused to convert the skeleton to a larger grid patch $D_{H'\!\times\! W'}$ where $H'\!>\! H,W'\!>\! W$. This process can be formally defined as

\begin{equation}{\label{eq_pls}}
\boldsymbol{Y}={\rm reshape}(\boldsymbol{\Phi}_2\cdot(\boldsymbol{\Lambda}_2\cdot(\boldsymbol{\Phi}\cdot((\boldsymbol{\Lambda}\cdot \boldsymbol{A})\cdot \boldsymbol{X})))),
\end{equation}

where $\boldsymbol{\Phi}\cdot((\boldsymbol{\Lambda}\cdot \boldsymbol{A})\cdot \boldsymbol{X})$ constructs the base grid path $D_{H\!\times\! W}$ using the first-stage transform pair of UPT and GIT as defined by \cref{eq_UPT} and \cref{eq_GIT}; $\boldsymbol{\Lambda}_2\!\in\! \mathbb{R}^{H'W'\!\times\! HW}$ is the second-stage UPT for up-sampling the grid patch from the base size of $H\!\times\! W$ to the target size of $H'\!\times\! W'$, and $\boldsymbol{\Phi}_2\!\in\! \mathbb{R}^{H'W'\!\times\! H'W'}$ is the second-stage GIT for allocating the up-sampled features to the target grid patch.

The transform pair of UPT and GIT learned in the first stage is fixed when learning the second-stage transform pair of UPT and GIT.
Furthermore, the Ske2Grid convolution network maintains the same structure during different training stages, and thus the network trained for the first-stage grid patch is also used as a pre-trained model to initialize the network training in the second stage.
These two aspects constitute the PLS, which can be naturally used in a cascaded way consisting of multiple transform pairs of UPT and GIT to boost the performance of the Ske2Grid convolution network. Thanks to the lightweight designs of UPT and GIT, they introduce negligible extra computation cost to the Ske2Grid convolution network during inference.
The cascaded structure of multiple UPT and GIT pairs facilitated by the PLS incorporates extra nodes interpolated from the original graph nodes and enables the regular convolution to capture various ordered topological relations, strengthening the representation learning ability of our grid patch.


\subsection{Network Construction}
\label{network}

With the above designs for grid representation learning, convolution operation in our Ske2Grid can model a learnable set of enriched grid cells for improved action representation.
The grid patch is fed into a Ske2Grid convolutional network for action recognition, as shown in \cref{ske2grid_pipeline}.
By jointly training the UPT, GIT and the Ske2Grid convolution network in an end-to-end manner facilitated with the PLS, the enriched topological connections on the skeleton can be automatically learned conditioned on a target action recognition dataset.
To make our Ske2Grid be applicable to popular GCN architectures without modifying their built-in temporal modules, we construct the Ske2Grid convolution network upon prevailing GCNs through simply replacing the spatial graph convolution with the convolution operation upon our learnable grid patch, without changing the structure of the backbone network and their temporal modules.

\section{Experiments}
\label{experiments}


\subsection{Datasets}
\label{dataset}

Six mainstream datasets are considered in the experiments.

\textbf{NTU-60} \cite{Ref00_Dataset_NTU} is the first large-scale multi-modality skeleton-based action recognition dataset. It contains 56,880 skeleton action sequences which are performed by 40 volunteers and categorized into 60 classes. There are two popular validation protocols for this dataset: cross-subject (XSub) and cross-view (XView).

\textbf{NTU-120} \cite{Ref00_Dataset_NTU120} is the extended version of NTU-60. It contains extra 57,367 skeleton sequences from 60 extra action classes. Two popular validation protocols for this dataset are: cross-subject(XSub) and cross-setup (XSet).

\textbf{FineGym99} \cite{Ref00_Dataset_finegym} is a newly released fine-grained action recognition dataset with 29,000 videos of 99 gymnastic action classes.
\textbf{HMDB51} \cite{Ref00_Dataset_hmdb51} and \textbf{UCF101} \cite{Ref00_Dataset_ucf101} are two early popular action recognition datasets collected from the web. HMDB51 contains 6,700 videos from 51 classes and UCF101 contains 13,000 videos from 101 classes.
\textbf{Diving48} \cite{Ref00_Dataset_diving48} contains over 18,000 video clips of competitive diving actions, spanning 48 fine-grained dive classes.

\textbf{Skeletons for all datasets.} Regarding the experiments using 2D estimated skeleton, we use the 2D poses on these datasets provided by PYSKL \cite{Ref05_Tools_pyskl} for fair comparison, which are detected by HRNet \cite{Ref08_Others_HRNet} pre-trained on COCO \cite{Ref08_Others_COCO}. While there're ground-truth (GT) human bounding boxes on FineGym99, the human detector for all the other datasets is Faster-RCNN \cite{Ref08_Others_Faster_RCNN} with ResNet50 \cite{Ref08_Others_ResNet} as backbone.
For the experiments on NTU-60 and NTU-120 using 3D GT skeleton, we also use the preprocessed 3D poses on these datasets provided by PYSKL for fair comparison.

\subsection{Implementation Details}
\label{implement}

Regarding skeleton input, we use 2D joint coordinates plus the estimated scores as joint features for 2D skeleton input, and use 3D joint coordinates for GT 3D skeleton input, and thus $C=3$ in both cases.
For the initialization of the skeleton to grid representation transforms, we use random initialization for $\boldsymbol{\Psi}$ in the GIT.
To better enjoy the regulation from the adjacency matrix in the UPT, $\boldsymbol{\Lambda}$ is initialized as an identity matrix of size $N\!\times\! N$ cascaded by a random matrix.

\begin{figure}[t]
\vskip 0.in
\begin{center}
\centerline{\includegraphics[scale=0.25]{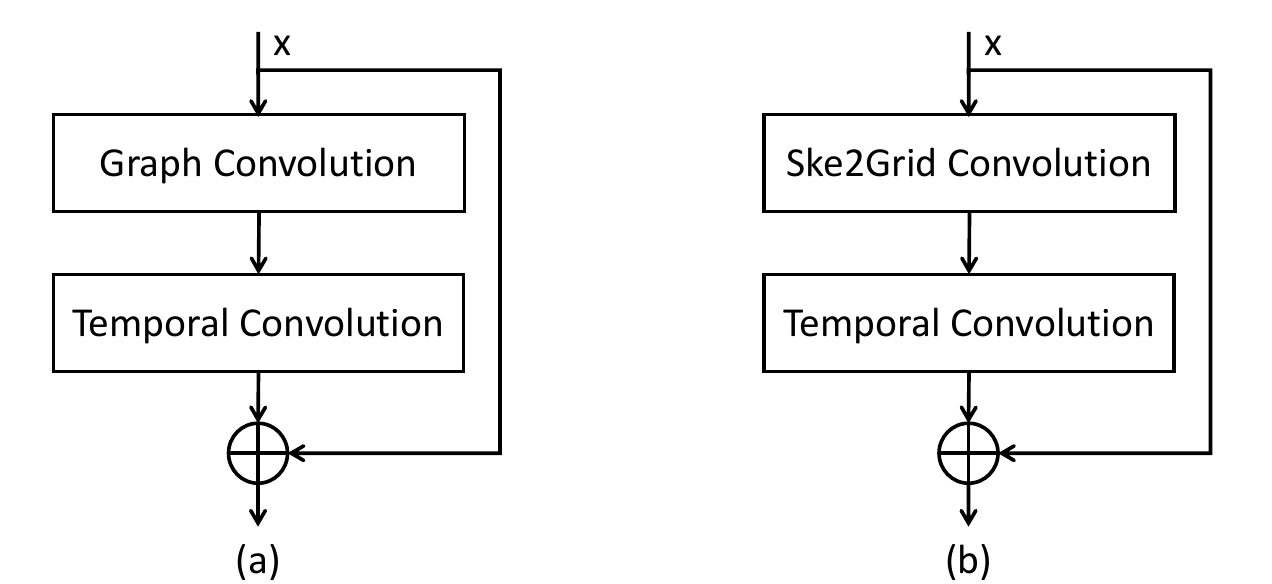}}
\caption{The difference between the ST-GCN block (a) and our ST-GCN$^{\star}$ block (b).}
\label{stgcn_block}
\end{center}
\vskip -0.4in
\end{figure}

Regarding network construction, we choose the prevailing ST-GCN \cite{Ref03_GCN_ST_GCN} as backbone network and modify the basic block through replacing the spatial graph convolution with regular 2D convolution upon our grid patch, as shown in \cref{stgcn_block}.
We denote this modified network as ST-GCN$^\star$.
The kernel size is $3\!\times\! 3$, unless otherwise stated.
The transform pair of UPT and GIT for $D_{H\!\times\! W}$ is added before the action recognition network, formulating our basic Ske2Grid$_{H\!\times\! W}$ framework.

For fair comparisons, we use the popular comprehensive skeleton-based action recognition toolbox PYSKL \cite{Ref05_Tools_pyskl} to implement all the experiments.
There are mainly two training settings.
For the experiments conducted to explore the effects of our core designs, we use the vanilla training strategy of PYSKL as in ST-GCN \cite{Ref03_GCN_ST_GCN} for fair and clean comparisons, in which each model is trained for $80$ epochs with the learning rate decayed by $10$ at $10^{th}$ and $50^{th}$ epochs respectively.
For the main experiments to explore the capability of Ske2Grid as shown in \cref{6 benchmarks},\cref{ntu_3d}, \cref{ablative_backbone} and \cref{compare_sota}, we use the latest common experimental setups in PYSKL, in which each model is trained for $80$ epochs with the cosine schedule of learning rate.
In both settings, the initial learning rate is set to $0.1$, the batch size is $128$, the momentum is set to $0.9$, the weight decay is $5\!\times\! 10^{-4}$, and the Nesterov momentum is used for the optimizer.
We report the validation top-1 accuracy in all the experiments except for \cref{compare_sota}.
When comparing with state-of-the-art methods, we also follow the common practice as in \citet{Ref02_CNN_PoseC3D} to report the testing top-1 accuracy for fair comparison, which is slightly better than the validation performance due to data augmentation.

\begin{table}[t]
\caption{Effects of the three key designs for learning the grid representation in our Ske2Grid using NTU-60 XSub benchmark.
}
\label{three designs}
\vskip 0.0in
\begin{center}
\begin{small}
\begin{sc}
\scalebox{0.825}{
\begin{tabular}{c|ccc|c}
\toprule
Method    & GIT & UPT & PLS & Top-1 Acc.($\%$) \\
\midrule
ST-GCN    & $-$ & $-$ & $-$ & 85.25 \\
\hline
\multirow{2}{*}{Ske2Grid$_{5\times 5}$}  & $\surd$ & $\times$ & $\times$ & 84.70 \\
  & $\surd$ & $\surd$ & $\times$ & \textbf{86.20} \\
\hline
\multirow{4}{*}{Ske2Grid$_{6\times 6}$}  & $\surd$ & $\times$ & $\times$ & 84.04 \\
  & $\surd$ & $\surd$ & $\times$ & 85.29 \\
  & $\surd$ & $\times$ & $\surd$ & 85.61 \\
  & $\surd$ & $\surd$ & $\surd$ & \textbf{87.87} \\
\hline
\multirow{4}{*}{Ske2Grid$_{7\times 7}$}  & $\surd$ & $\times$ & $\times$ & 85.93 \\
  & $\surd$ & $\surd$ & $\times$ & 86.35 \\
  & $\surd$ & $\times$ & $\surd$ & 85.96 \\
  & $\surd$ & $\surd$ & $\surd$ & \textbf{88.26} \\
\hline
\multirow{4}{*}{Ske2Grid$_{8\times 8}$}  & $\surd$ & $\times$ & $\times$ & 84.81 \\
  & $\surd$ & $\surd$ & $\times$ & 85.81 \\
  & $\surd$ & $\times$ & $\surd$ & 86.11 \\
  & $\surd$ & $\surd$ & $\surd$ & \textbf{88.55} \\
\hline
\multirow{4}{*}{Ske2Grid$_{9\times 9}$}  & $\surd$ & $\times$ & $\times$ & 84.10 \\
 & $\surd$ & $\surd$ & $\times$ & 84.95 \\
 & $\surd$ & $\times$ & $\surd$ & 86.60 \\
 & $\surd$ & $\surd$ & $\surd$ & \textbf{88.37} \\
\bottomrule
\end{tabular}}
\end{sc}
\end{small}
\end{center}
\vskip -0.2in
\end{table}

\subsection{Effects of Core Designs in Ske2Grid}
\label{core design}

We first conduct experiments to analyze the effects of our three core designs for learning our grid representation, including the UPT, GIT and PLS.
We compare the performance of ST-GCN$^\star$ using different combinations of these three designs for learning the grid patch, as shown in \cref{three designs}. The ``PLS'' setting indicates training the Ske2Grid$_{H\!\times\! W}$ progressively from Ske2Grid$_{(H\!-\!1)\!\times\! (W\!-\!1)}$ in a cascaded manner.

Ske2Grid of ``GIT-only'' fills the grid cells with all the distinct graph nodes and the remaining grid cells without the repetitive constraint. As shown, there is no obvious correlation between the performance and grid patch size. Together with ``PLS'', the performance improves progressively with the increase of the grid patch size, but the performance gain is not significant. This is probably due to the uncertainty brought by the redundant grid cells.

With the ``GIT \& UPT'' setting, the performance of Ske2Grid becomes stably improved over the baseline. However, when the grid patch grows to $D_{9\!\times\! 9}$, the performance drops below the baseline. In this case, the number of grid cells is about $5$ times the number of joints in the skeleton input. It becomes challenging to interpolate the skeleton graph for satisfactory expressiveness using only one-step UPT.

Combining all the three designs, our Ske2Grid consistently outperforms the baseline with significant performance margin. And the performance improves progressively with the increase of the grid patch size, yielding $3.3\%$ top-1 margin when using $D_{8\!\times\! 8}$, well demonstrating the effectiveness of our Ske2Grid in learning improved skeleton features for action recognition.
In the following main experiments, we train our Ske2Grid models using all the three designs.
Considering the training overhead, we use the progressive learning strategy cascaded up to $3$ times (4 stages), e.g. from $D_{5\!\times\!5}$ to $D_{8\!\times\!8}$ by a step of $1$ for 2D estimated skeleton input.

\begin{table}[t]
\caption{Main results of our Ske2Grid on six datasets using 2D estimated skeletons. We report the performance of our Ske2Grid$_{8\!\times\! 8}$.
}
\label{6 benchmarks}
\vskip 0.0in
\begin{center}
\begin{small}
\begin{sc}
\scalebox{0.825}{
\begin{tabular}{c|c|c|c|c}
\toprule
Method    & \multicolumn{2}{|c|}{Dataset} & Top-1 Acc.($\%$) & $\Delta$Top-1($\%$) \\
\midrule
ST-GCN     & \multirow{4}{*}{NTU-60} &  \multirow{2}{*}{XSub} & 88.23 & \multirow{2}{*}{\textbf{+3.70}}\\
Ours       & & & \textbf{91.93} & \\
\cline{1-1} \cline{3-5}
ST-GCN     & & \multirow{2}{*}{XView} & 96.63 & \multirow{2}{*}{\textbf{+1.12}}\\
Ours       & & &  \textbf{97.75} & \\
\hline
ST-GCN     & \multirow{4}{*}{NTU-120}  &  \multirow{2}{*}{XSub} & 83.56 & \multirow{2}{*}{\textbf{+1.24}}\\
Ours       & & & \textbf{84.80} & \\
\cline{1-1} \cline{3-5}
ST-GCN     & & \multirow{2}{*}{XSet} & 83.84 & \multirow{2}{*}{\textbf{+3.69}}\\
Ours       & & &  \textbf{87.53} & \\
\hline
ST-GCN     & \multicolumn{2}{c|}{\multirow{2}{*}{FineGym99}} & 91.17 & \multirow{2}{*}{\textbf{+0.65}}\\
Ours       & \multicolumn{2}{c|}{} &  \textbf{91.82} & \\
\hline
ST-GCN     & \multicolumn{2}{|c|}{\multirow{2}{*}{UCF101}}  & 69.23 & \multirow{2}{*}{\textbf{+3.83}}\\
Ours       & \multicolumn{2}{c|}{} & \textbf{73.06} & \\
\hline
ST-GCN     & \multicolumn{2}{|c|}{\multirow{2}{*}{HMDB51}}  & 47.25 & \multirow{2}{*}{\textbf{+1.12}}\\
Ours       & \multicolumn{2}{c|}{} & \textbf{48.37} & \\
\hline
ST-GCN     & \multicolumn{2}{c|}{\multirow{2}{*}{Diving48}}  & 38.32 & \multirow{2}{*}{\textbf{+5.79}}\\
Ours       & \multicolumn{2}{c|}{} & \textbf{44.11} & \\
\bottomrule
\end{tabular}}
\end{sc}
\end{small}
\end{center}
\vskip -0.2in
\end{table}

\begin{table}[t]
\caption{Performance comparison of our Ske2Grid using 3D GT skeleton inputs. We report the performance of Ske2Grid$_{9\!\times\! 9}$.
}
\label{ntu_3d}
\vskip 0.0in
\begin{center}
\begin{small}
\begin{sc}
\scalebox{0.825}{
\begin{tabular}{c|c|c|c|c}
\toprule
Method    & \multicolumn{2}{|c|}{Dataset} & Top-1 Acc.($\%$) & $\Delta$Top-1($\%$) \\
\midrule
ST-GCN     & \multirow{4}{*}{NTU-60} & \multirow{2}{*}{XSub} & 87.62 & \multirow{2}{*}{\textbf{+0.63}}\\
Ours       & & & \textbf{88.25} & \\
\cline{1-1} \cline{3-5}
ST-GCN     & & \multirow{2}{*}{XView}  & 95.08 & \multirow{2}{*}{\textbf{+0.64}}\\
Ours       & & & \textbf{95.72} & \\
\hline
ST-GCN     & \multirow{4}{*}{NTU-120} & \multirow{2}{*}{XSub}  & 81.13 & \multirow{2}{*}{\textbf{+1.60}}\\
Ours       & & & \textbf{82.73} & \\
\cline{1-1} \cline{3-5}
ST-GCN     & & \multirow{2}{*}{XSet}  & 83.63 & \multirow{2}{*}{\textbf{+1.44}}\\
Ours       & & & \textbf{85.07} & \\
\bottomrule
\end{tabular}}
\end{sc}
\end{small}
\end{center}
\vskip -0.2in
\end{table}

\subsection{Main Results}
\label{main results}

\cref{6 benchmarks} shows the main results of our Ske2Grid on six action recognition datasets using $D_{8\!\times\!8}$.
As can be seen, our method consistently outperforms the baseline on all datasets, showing its good generalization ability for improved skeleton-based action recognition.
The performance margin on FineGym99 is relatively small among these datasets.
This maybe because the 2D estimated skeleton on this dataset is based on GT human bounding boxes which is more accurate, compared to the other datasets.
The performance improvement from our grid representation becomes marginal.
This conjecture is further verified by the large top-1 accuracy margin on Diving48.
Since it contains challenging fine-grained dive classes and relatively inaccurate 2D estimated skeleton benefits the most from the enriched grid representation to recognize the fine-grained actions.

\begin{table}[t]
\caption{Impacts of the convolution operation in our Ske2Grid using NTU-60 XSub benchmark. We use ST-GCN to directly model our grid patch obtained by the UPT and GIT. ST-GCN$^{\star}$ is with the convolution defined in our Ske2Grid. ``${\diamond}$'' denotes that the transform is well-learned and fixed during training.
}
\label{ablative_network}
\vskip 0.in
\begin{center}
\begin{small}
\begin{sc}
\scalebox{0.85}{
\begin{tabular}{c|c|c|c|c}
\toprule
Network  &  UPT & GIT & $D$ size & Top-1 Acc.($\%$) \\
\midrule
ST-GCN  & $-$ & $-$ & $-$  &  85.25 \\
\hline
\multirow{4}{*}{ST-GCN} & $\surd$ & $\surd$  & \multirow{5}{*}{$5\times 5$}  & 80.39 \\
 & $\diamond$   & $\surd$       &   & 82.14 \\
 & $\surd$      & $\diamond$    &   & 81.94 \\
 & ${\diamond}$ & ${\diamond}$  &   & 84.34 \\
\cline{1-3} \cline{5-5}
ST-GCN$^{\star}$ & $\surd$ & $\surd$  &    & \textbf{86.20} \\
\midrule
\multirow{4}{*}{ST-GCN} & $\surd$ & $\surd$  & \multirow{5}{*}{$6\times 6$}   & 80.63 \\
 & ${\diamond}$  & $\surd$        &   & 79.71 \\
 & $\surd$       & ${\diamond}$   &   & 81.26 \\
 & ${\diamond}$  & ${\diamond}$   &   & 84.53 \\
\cline{1-3} \cline{5-5}
ST-GCN$^{\star}$ & $\surd$ & $\surd$ &  & \textbf{87.87} \\
\bottomrule
\end{tabular}}
\end{sc}
\end{small}
\end{center}
\vskip -0.2in
\end{table}

\subsection{Ablative Studies}
\label{ablation}

We further provide a number of ablative studies to have a deep analysis of our Ske2Grid.

\textbf{Ske2Grid using 3D skeleton input.} In addition to 2D estimated skeleton input, we use 3D GT skeleton consisting of $25$ keypoints on NTU-60 and NTU-120 as input to conduct experiments as shown in \cref{ntu_3d}.
Our Ske2Grid is trained from $D_{6\!\times\!6}$ to $D_{9\!\times\!9}$ by a step of $1$.
As shown, our method outperforms the baseline under different validation protocols on both datasets, demonstrating the strong generalization capability of our Ske2Grid for improved motion-captured skeleton-based action recognition.

\textbf{Impacts of the convolution operation.} To analyze the impacts of the convolution operation in our Ske2Grid, we use graph convolution to directly model our grid representation through treating the grid patch as a regular graph with vertical and horizontal connections between adjacent grid cells.
The comparison is shown in \cref{ablative_network}.

Firstly, ST-GCN is used to model a grid patch with both the UPT and GIT being learnable.
Not surprisingly, the performance drops significantly. As discussed before, graph convolution convolves each grid cell with their adjacent grid cells using its specific kernels. As the layout of our grid patch is learned with the network, it is difficult to adapt the graph convolution for modeling the changing topological relationship of grid cells during training.

Further, we reuse the UPT and GIT well learned by our Ske2Grid to train ST-GCN for modeling the grid patch.
As shown, the performance is relatively good when fixing both the UPT and GIT, illustrating the capability of graph convolution in modeling fixed graph. However the fixed regular layout of grid patch still limits the expressiveness of skeleton for action representation.
ST-GCN$^{\star}$ with learnable grid patch achieves the best performance under different size settings, demonstrating the effectiveness of our convolution operation in modeling the feature interactions among a learnable set of grid cells for improved action recognition.

\begin{table}[t]
\caption{Performance ($\%$) comparison of different base grid patches and progressive steps for the PLS setting in our Ske2Grid using NTU-60 XSub benchmark. $D_K$ is short for $D_{K\!\times\! K}$.
}
\label{ablative_step}
\vskip 0.0in
\begin{center}
\begin{small}
\begin{sc}
\scalebox{0.85}{
\begin{tabular}{c|c|c|c|c|c|c}
\toprule
Start & PLS$_1$ & Acc. & PLS$_2$ & Acc. & PLS$_3$ & Acc.\\
\midrule
\multirow{3}{*}{$D_{5}$} &  $D_{6}$  &  87.87&  $D_{7}$  &  88.26&  $D_{8}$  &  \textbf{88.55}\\
   &  $D_{7}$  & 87.98 &  $D_{9}$  & 88.15   &  $D_{11}$  & 88.47 \\
   &  $D_{8}$  & 87.92 &  $D_{11}$  & 87.82   &  $-$  & $-$ \\
\hline
\multirow{3}{*}{$D_{6}$} &  $D_{7}$   &  86.27 &  $D_{8}$  &  86.41&  $D_{9}$  &  87.57\\
   &  $D_{8}$  & 86.23  &  $D_{10}$  & 87.69   &  $D_{12}$  & 88.49 \\
   &  $D_{9}$  & 86.44  &  $D_{12}$  & 87.59   &  $-$  & $-$ \\
\hline
\multirow{3}{*}{$D_{7}$} &  $D_{8}$  &  87.15  &  $D_{9}$  &  87.30 &  $D_{10}$  &  87.72\\
   &  $D_{9}$  & 86.85  &  $D_{11}$  & 87.18   &  $D_{13}$  & 87.20 \\
   &  $D_{10}$  & 87.17  &  $D_{13}$  & 87.59   &  $-$  & $-$ \\
\bottomrule
\end{tabular}}
\end{sc}
\end{small}
\end{center}
\vskip -0.2in
\end{table}

\textbf{Impacts of different PLS settings.}
We exhaustively explore the starting size of grid patch and the progressive step in the PLS, as shown in \cref{ablative_step}.
The performance increases progressively under all different settings.
As shown, training our Ske2Grid from a small grid patch with the number of grid cells being close to the number of joints in the skeleton input progressively with a step of $1$ performs better.

\begin{table}[t]
\caption{Performance comparison when using other grid patch sizes in our Ske2Grid using NTU-60 XSub benchmark.
}
\label{ablative_rect}
\vskip 0.0in
\begin{center}
\begin{small}
\begin{sc}
\scalebox{0.85}{
\begin{tabular}{c|c}
\toprule
Grid Patch &  Top-1 Acc.($\%$) \\
\midrule
$D_{5\times5}$ &    \textbf{86.20} \\
$D_{4\times6}$ &    85.61 \\
$D_{4\times7}$ &    84.64 \\
$D_{5\times6}$ &    85.91 \\
$D_{5\times7}$ &    84.25 \\
\bottomrule
\end{tabular}}
\end{sc}
\end{small}
\end{center}
\vskip -0.2in
\end{table}

\begin{table*}[t]
\vskip 0.in
\caption{Performance ($\%$) comparison of our Ske2Grid with state-of-the-art methods on NTU-60 and NTU-120 using estimated 2D skeletons. We report the performance of Ske2Grid$_{8\!\times\!8}$. The ``$J$'', ``$B$'', ``$JM$'' and ``$BM$'' represent the joint, bone, joint motion and bone motion data modalities, respectively. The ``$\diamond$'' denotes using the same human skeletons as that provided by \citet{Ref05_Tools_pyskl}. The ``${\ast}$'' represents the average model fusion of our Ske2Grid from $D_{5\!\times\!5}$ to $D_{8\!\times\!8}$. Best results are bolded.
}
\label{compare_sota}
\vskip -0.2in
\begin{center}
\begin{small}
\begin{sc}
\scalebox{0.85}{
\begin{tabular}{c|cccc}
\toprule
Method    & N60-XSub & N60-XView & N120-XSub & N120-XSet \\
\midrule
2S-AGCN \cite{Ref03_GCN_Two_Stream_AGCN} ($J$+$B$) & 88.5 & 95.1 & $-$ & $-$ \\
MS-AAGCN \cite{Ref03_GCN_MS_AAGCN} ($J$+$B$+$JM$+$BM$) & 90.0 & 96.2 & $-$ & $-$ \\
AS-GCN \cite{Ref03_GCN_AS_GCN} ($J$)  & 86.8 & 94.2 & 78.3 & 79.8 \\
CA-GCN \cite{Ref03_GCN_Context_Aware_GCN} ($J$) & 83.5 & 91.4 & $-$ & $-$ \\
Shift-GCN \cite{Ref03_GCN_Shift_GCN}($J$+$B$+$JM$+$BM$)  & 90.7 & 96.5 & 85.9 & 87.6 \\
MS-G3D \cite{Ref03_GCN_MS_G3D}($J$+$B$)  & 91.5 & 96.2 & 86.9 & 88.4 \\
CTR-GCN \cite{Ref03_GCN_CTR-GCN}($J$+$B$+$JM$+$BM$)  & 92.4 & 96.8 & \textbf{88.9} & 90.6 \\
\midrule
HyperGNN \cite{HyperGNN}($J$+$B$+$JM$+$BM$)  & 89.5 & 95.7 & $-$ & $-$ \\
Sybio-GNN \cite{SybioGNN}($J$+$B$)  & 90.1 & 96.4 & $-$ & $-$ \\
STST \cite{STST}($J$+$B$+$JM$+$BM$)  & 91.9 & 96.8 & $-$ & $-$ \\
ST-TR \cite{ST-TR}($J$)  & 88.7 & 95.6 & 81.9 & 84.1 \\
Hyperformer \cite{Hyperformer}($J$)  & 90.7 & 95.1 & 86.5 & 88.1 \\
\midrule
PoseConv3D  \cite{Ref02_CNN_PoseC3D}($J$) $\diamond$ & 93.7 & 96.6 & 86.0 & 89.6 \\
\midrule
MS-G3D ($J$+$B$) $\diamond$  & 92.2 & 96.6 & 87.2 & 89.0 \\
ST-GCN ($J$) $\diamond$ & 88.9 & 96.8 & 84.0 & 84.1 \\
\midrule
Ske2Grid ($J$) $\diamond$  & 92.3 & 97.9 & 85.3 & 87.9 \\
Ske2Grid$\ast$ ($J$) $\diamond$  & 93.0  & 98.2 & 85.9 & 88.7 \\
Ske2Grid$\ast$ ($J$+$B$+$JM$+$BM$) $\diamond$  & \textbf{93.8}  & \textbf{98.6} & 87.3 & \textbf{90.8} \\
\bottomrule
\end{tabular}}
\end{sc}
\end{small}
\end{center}
\vskip -0.2in
\end{table*}

\begin{figure}[t]
\vskip 0.in
\begin{center}
\centerline{\includegraphics[scale=0.25]{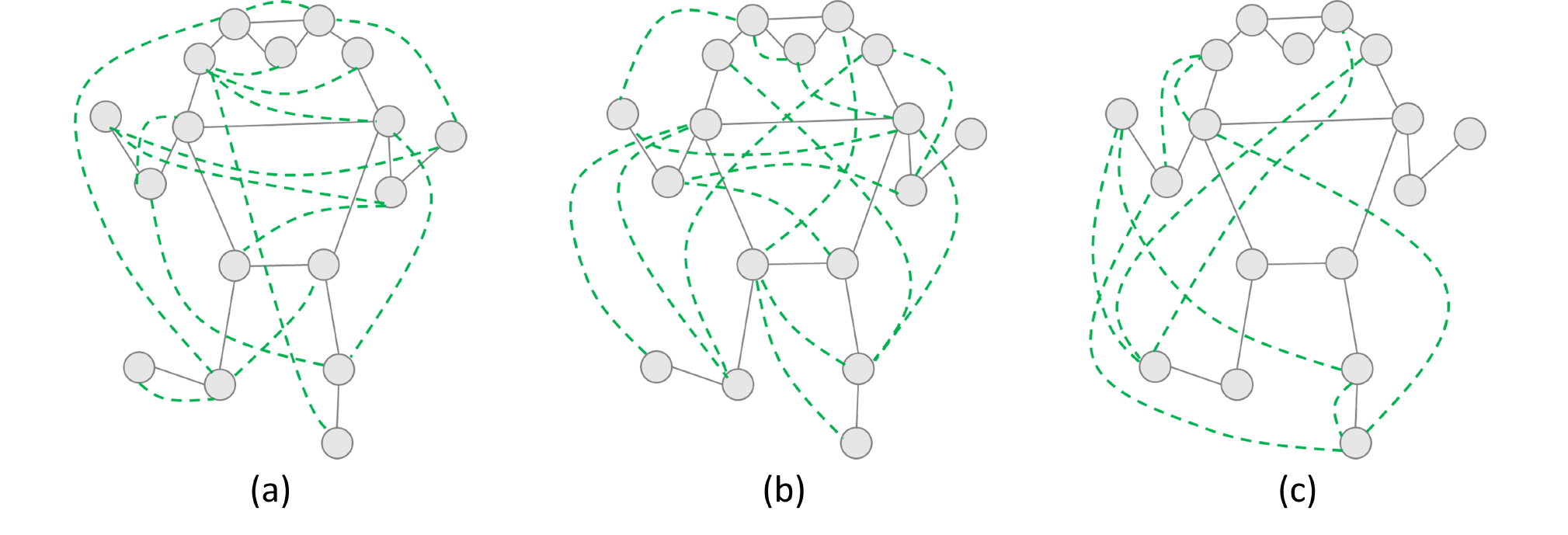}}
\vskip -0.15in
\caption{Visualization of the learnt topological relations on the skeleton graph reflected by (a) $D_{5\!\times\!5}$; (b) $D_{6\!\times\!6}$; and (c) $D_{7\!\times\!7}$.}
\label{vis}
\end{center}
\vskip -0.3in
\end{figure}

\begin{table}[t]
\caption{Performance ($\%$) comparison when constructing a network upon CTR-GCN \cite{Ref03_GCN_CTR-GCN} by our Ske2Grid using 3D GT skeletons. We report the performance of Ske2Grid$_{9\!\times\! 9}$.
}
\label{ablative_backbone}
\vskip 0.0in
\begin{center}
\begin{small}
\begin{sc}
\scalebox{0.85}{
\begin{tabular}{c|c|c|c|c}
\toprule
\multirow{2}{*}{Method} & \multicolumn{2}{|c|}{NTU-60} & \multicolumn{2}{|c}{NTU-120}\\
\cline{2-5}
 & XSub & XView & XSub & XSet\\
\midrule
CTR-GCN   &  89.62  & 94.83  &  84.57 & 85.95 \\
\midrule
Ours   &  \textbf{90.08}  &  \textbf{95.09}  &  \textbf{84.79} & \textbf{86.04} \\
\bottomrule
\end{tabular}}
\end{sc}
\end{small}
\end{center}
\vskip -0.2in
\end{table}

\textbf{Ske2Grid using other grid patch sizes.}
\cref{ablative_rect} shows the performance of our Ske2Grid when using different grid patch sizes without the PLS setting.
We can see that the performance of using squared grid patches is slightly better than that of using rectangular grid patches. Therefore, we use squared grid patches as default setting in the experiments.

\textbf{Ske2Grid applying to other GCNs.}
We construct a network upon CTR-GCN \cite{Ref03_GCN_CTR-GCN} through replacing the graph convolution in the basic block of spatial modeling with our convolution operation, and the performance is shown in \cref{ablative_backbone}.
Although CTR-GCN is a recent advanced GCN-based solution, our Ske2Grid still achieves acceptable performance gains under the same benchmark settings.

\textbf{Visualization of the learnt layouts in Ske2Grid.}
The grid patch in Ske2Grid is filled by the up-sampled graph nodes, and its layout reflects the topological relations of joints on the skeleton graph.
We visualize the learnt connections among graph nodes on the skeleton using the layouts of progressively learnt grid patches in Ske2Grid$_{8\!\times\!8}$, as shown in \cref{vis}, illustrating the capability of our method for modeling learnable skeleton feature interactions.

\textbf{Comparison of Ske2Grid with state-of-the-art methods.}
\cref{compare_sota} shows a performance comparison of our Ske2Grid with a lot of state-of-the-art methods on NTU-60 and NTU-120.
We collect the best results reported in the original papers of all the methods shown above PoseConv3D \cite{Ref02_CNN_pa3d}.
Prevailing GCN-based methods (the upper 7 solutions in \cref{compare_sota}) improve ST-GCN through constructing skeleton graph with dynamic topologies and incorporating extra contextual information to increase the receptive field of graph convolution, as we discussed in \cref{related_works}.
Recently proposed hypergraph and transformer based methods (the middle 5 solutions in \cref{compare_sota}) either introduce local and global hyperedges to encode higher-order feature dependencies, or treat each joint of the human body as a token and use spatial and temporal self-attention operators to capture feature dependencies.
As shown, our Ske2Grid with single ``joint'' modality outperforms or achieves comparable performance with these existing solutions.
PoseConv3D \cite{Ref02_CNN_pa3d} takes the stacked heatmaps (with the spatial size of $56\!\times\! 56$) of skeleton joints obtained by 2D pose estimators as input and utilizes computationally intensive 3D CNN to recognize actions.
Using much more compact grid representation (spatial size of $8\!\times\! 8$) and lightweight 2D convolution network, our Ske2Grid achieves comparable performance with PoseConv3D, which is promising.
Taking the advantage of our Ske2Grid with the progressive training, we ensemble the four models from $D_{5\!\times\!5}$ to $D_{8\!\times\!8}$ and obtain a significant performance boost, showing the complementarity among different grid patch scales.
When further combining with the bone, joint motion and bone motion modalities commonly used in other methods, our Ske2Grid performs the best on three out of four benchmarks.

\section{Conclusions}
\label{conclusion}

This paper presents Ske2Grid, a new representation learning framework for improved skeleton-based action recognition.
In Ske2Grid, we define a regular convolution operation upon a compact image-like grid patch constructed and learned through three novel designs namely UPT, GIT and PLS.
The layout of the grid representation is learned with the Ske2Grid convolution network, and convolution operation upon the grid patch models a learnable set of grid cells, which improves the modeling of feature interactions among active coordinated human joints for effective action recognition.
The efficacy of our Ske2Grid is well validated by thorough experiments on six public benchmarks.


%
%


\bibliography{skeleton_action}
\bibliographystyle{icml2023}



\end{document}